\pdfoutput=1
\documentclass{article}

\PassOptionsToPackage{numbers,sort&compress}{natbib}
\usepackage[preprint]{neurips_2026}

\usepackage[utf8]{inputenc}
\usepackage[T1]{fontenc}
\usepackage{url}
\usepackage{booktabs}
\usepackage{amsfonts}
\usepackage{amssymb}
\usepackage{array}
\newcolumntype{L}[1]{>{\raggedright\arraybackslash}p{#1}}
\usepackage{longtable}
\usepackage{amsmath}
\usepackage{nicefrac}
\usepackage{microtype}
\usepackage{xcolor}
\usepackage{graphicx}
\usepackage{float}
\usepackage{algorithm}
\usepackage{algpseudocode}
\usepackage{enumitem}

\numberwithin{figure}{section}
\numberwithin{table}{section}

\setlist[itemize]{itemsep=1pt,topsep=3pt}
\setlist[enumerate]{itemsep=1pt,topsep=3pt}

\usepackage[hidelinks,colorlinks=true,citecolor=blue,linkcolor=blue,urlcolor=blue]{hyperref}

\title{OR-Transformer: Scaling Real-Time Decision-Making to 1,000 Items}

\author{%
  Shuze Daniel Liu$^{1,2}$ \quad David Simchi-Levi$^{1,2}$ \quad Claire Chen$^{3}$ \\
  \bfseries Chutong Gao$^{1,2}$ \quad Shangtong Zhang$^{4}$ \\[0.2em]
  $^{1}$Massachusetts Institute of Technology \quad
  $^{2}$Purdue University \\
  $^{3}$California Institute of Technology \quad
  $^{4}$University of Virginia \\
}

\begin{document}

\maketitle

\begin{abstract}
Modern supply chain operations can require coordinating replenishment across thousands of heterogeneous items under correlated stochastic demand, heterogeneous lead times, and shared fixed ordering costs, yielding observation spaces exceeding $10^4$ dimensions. At this scale, rolling-horizon stochastic mixed-integer linear programs (MILPs) become prohibitively slow, while standard reinforcement learning (RL) methods face increasingly challenging credit assignment in high-dimensional action spaces. 
We introduce \textbf{OR-Transformer}, a deep reinforcement learning framework for joint replenishment under stochastic demand, with an item-permutation-equivariant Transformer architecture and pathwise-gradient training through the inventory dynamics. 
Across problem sizes up to 1,024 inventory items, OR-Transformer increasingly outperforms learning-based and rolling-horizon MILP baselines as scale grows. It also reduces online decision-making time by over 4 million times relative to MILP solvers, enabling real-time, large-scale deep RL in supply chain operations.

\end{abstract}

\section{Introduction}
Inventory management is a fundamental decision problem in retail and supply chain operations, concerning when to place orders and how much of each item to order over time. The economic consequences of these decisions can be substantial. For example, IHL Group estimates that out-of-stocks and overstocks together cost retailers approximately \$1.7 trillion annually, corresponding to about 6.2\% of global retail sales \citep{ihl2026inventory_report}. 
\begin{figure}[H]
  \centering
  \includegraphics[width=0.95\textwidth]{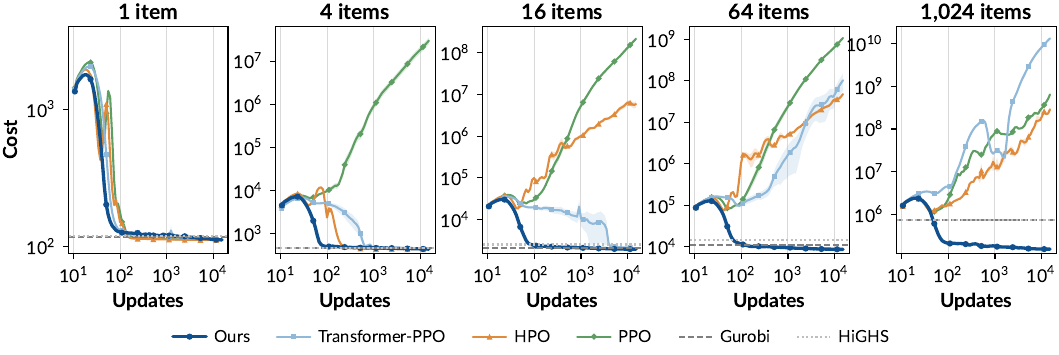}
  \vspace{-1em}
\caption{\textbf{Scaling performance from 1 to 1,024 items.}
OR-Transformer increasingly outperforms learning-based and rolling-horizon MILP baselines as problem size grows; lower cost is better.}
  \label{fig:line_plots_row}
\end{figure}
At large retailers, replenishment decisions are often coordinated across multiple products. For example, \citep{liu2023efficient_mathematics} reports that large supermarket chains such as Wal-Mart and Yonghui use joint replenishment strategies to reduce shipping costs. In such settings, a shared fixed ordering cost links the decisions across products: each joint order incurs this cost, and the decision maker must choose which products to include and in what quantities. This coupling gives rise to the \emph{joint replenishment problem} (JRP), a classical problem in operations research (OR) \citep{goyal1974determination_management_science,khouja2008review_ejor}. 

Scaling stochastic joint replenishment is difficult for both optimization and learning. With more than 1,000 items \citep{zhang2024wims_iclr}, observation and action spaces in our setting exceed \(10^4\) and \(10^3\) dimensions, respectively, making rolling-horizon stochastic MILPs prohibitively slow \citep{xiang2019rs_arxiv,barati2026gym_arxiv} and RL credit assignment increasingly difficult. Heterogeneous lead times and correlated demand introduce additional dependencies across time and items \citep{madeka2022deep_arxiv,chen2023robust_msom}. Moreover, the shared fixed ordering cost creates a discontinuity between zero and any positive order, challenging standard continuous-action RL methods \citep{schulman2017proximal_arxiv,haarnoja2018soft_icml,alvo2026policy_arxiv}. 

Prior work has studied learning-based joint replenishment, differentiable inventory control, and Transformer-based inventory policies, but existing approaches either omit the shared fixed-cost coupling considered here or study it at substantially smaller action dimensions \citep{vanvuchelen2020ppo_computers_industry,madeka2022deep_arxiv,alvo2026policy_arxiv,liu2026inventory_transformer_service_science}. More broadly, recent work shows that differentiable-simulator methods and other DRL methods can benefit substantially from inventory-structured policy parameterizations \cite{xie2026deepstock}, consistent with theoretical results motivating learning within structured inventory-policy classes \cite{xie2024vc}. A related line of work also explores how OR methods, LLM agents, and human decision-makers can be combined for inventory control \cite{baek2026ai}.

We introduce \textbf{OR-Transformer}, a deep RL framework for stochastic joint replenishment. It uses an item-permutation-equivariant Transformer to allow arbitrary reordering of inventory items, while self-attention captures dependencies across item states. The training method uses differentiable inventory dynamics to propagate pathwise gradients directly to continuous order quantities, enabling direct credit assignment in high-dimensional action spaces.

We evaluate OR-Transformer against learning-based baselines spanning multiple model architectures and training algorithms, as well as classical rolling-horizon stochastic MILP baselines \citep{xiang2019rs_arxiv,barati2026gym_arxiv}.  Our experiments demonstrate that OR-Transformer increasingly outperforms the baselines as problem size grows; see Figure \ref{fig:line_plots_row}. At 1,024 inventory items, OR-Transformer achieves approximately 0.35M discounted cost, reducing cost by 74.7\%--90.8\% relative to the learning-based baselines and by about 96\% relative to rolling-horizon MILP controllers despite allowing up to 10 minutes of optimization per decision. In runtime comparisons, OR-Transformer reduces online decision-making time by more than 4 million times relative to MILP solvers, offering a practical path toward real-time inventory control at large scale.

\section{Problem Formulation}
\label{sec:problem}

We consider a stochastic joint replenishment problem with $n$ items.
Each item $i$ has holding cost $h_i$, backlog cost $b_i$, lead time $L_i$,
and maximum order quantity $\bar q_i$. Placing any joint order incurs a
shared fixed cost $K$, and unmet demand is backlogged.
Let \(D_{i,t}\) denote the stochastic demand for item \(i\) at time \(t\). We model demands across items as correlated through an observed common stochastic factor \(F_t \in \mathbb R \). The full demand specification and additional formulation details are provided in Appendix~\ref{app:problem_formulation}.

Let $\mathbf I_t$ denote the vector of net inventories, $\mathbf P_t$ the outstanding orders scheduled to arrive in future time steps, and $\Theta$ the time-invariant parameters defining the problem instance. The state is $s_t=(\mathbf I_t,\mathbf P_t,F_t;\Theta)$. At each time step $t\in \mathbb N$, let $Y_t\in\{0,1\}$ indicate whether a joint order is
placed and let $\mathbf Q_t=(Q_{1,t},\ldots,Q_{n,t})^\top$ denote the
order quantities, with $0\le Q_{i,t}\le\bar q_i$. The action is $a_t=(Y_t,\mathbf Q_t)$. Orders for item \(i\) arrive after lead time \(L_i\), with outstanding orders tracked in \(\mathbf P_t\).

Let $[x]^+ \doteq \max\{x,0\}$. The one-step cost is
$C_t = K Y_t + \sum_{i=1}^{n}
\bigl(
h_i[I_{i,t}-D_{i,t}]^+
+
b_i[D_{i,t}-I_{i,t}]^+
\bigr)$.

Let $\pi$ denote the replenishment policy and $\gamma\in(0,1)$ the discount
factor. The objective is to minimize the expected infinite-horizon discounted
cost $\min_{\pi}\;
\mathbb E_{\pi}\!\left[
\sum_{t=0}^{\infty}\gamma^t C_t
\right].$
\section{Method}
\label{sec:method}

\begin{figure*}[t]
  \centering
  \includegraphics[width=0.95\textwidth]{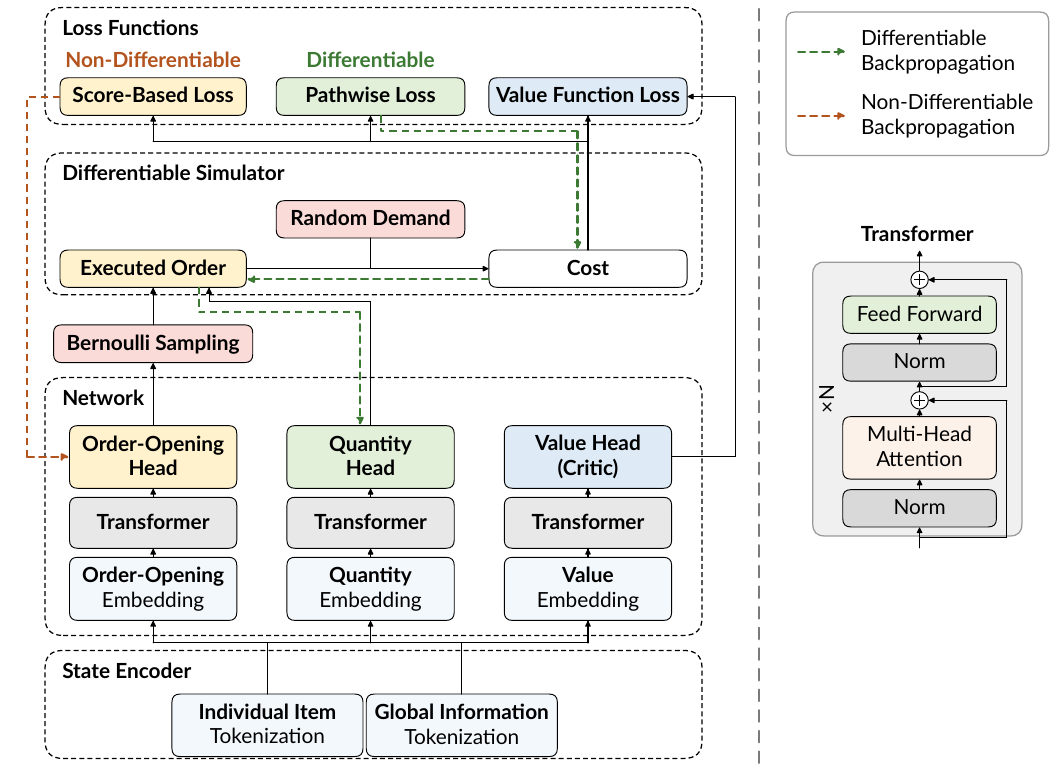}
  \caption{\textbf{OR-Transformer architecture and training.}
Independent Transformer networks produce the order-opening decision, item
quantities, and value estimate. Score-based gradients train the discrete
opening decision, while pathwise gradients propagate through the inventory
dynamics to train continuous quantities.}
\label{fig:or_transformer_architecture}
\end{figure*}

OR-Transformer consists of an item-permutation-equivariant Transformer policy and a
training procedure that propagates gradients through the inventory dynamics.
Figure~\ref{fig:or_transformer_architecture} summarizes the architecture and
gradient flow. Full details are provided in Appendix~\ref{app:detailed_method}.

\paragraph{Item-permutation-equivariant Transformer.}
The state is encoded into one global token and one token for each inventory
item. Three independent Transformer networks parameterize the order-opening
decision, the order quantities, and a critic that estimates expected discounted future cost. No positional or item-index embeddings are used, so arbitrarily reordering the inventory items
leaves the global outputs unchanged and reorders the quantity outputs
accordingly. Self-attention allows each item representation to incorporate information from other item states, capturing cross-item dependencies.

Let $p_t$ denote the probability of opening an order and 
$Q_{i,t}$ the quantity for item $i$. The policy outputs
\begin{equation}
Y_t\sim\operatorname{Bernoulli}(p_t),
\qquad
Q_{i,t}=\bar q_i\,\operatorname{sigmoid}
\!\left(\mathbf w_Q^\top\mathbf h_{i,t}^{Q}+c_Q\right),
\label{eq:policy_outputs_main}
\end{equation}
where \(\mathbf h_{i,t}^Q\) is the final Transformer representation of item
\(i\), \(\mathbf w_Q\) is the learned weight vector of the shared
quantity-output head, and \(c_Q\) is its learned scalar bias; and we recall that \(\bar{q}_i\) is the maximum order quantity for item $i$.

Writing \(p(s)\), \(\mathbf Q(s)\), and \(V(s)\) for the order-opening probability, quantity vector, and value estimate, respectively, for any item-permutation matrix \(\mathbf P\),
\begin{equation}
p(\mathbf P s)=p(s),\qquad
V(\mathbf P s)=V(s),\qquad
\mathbf Q(\mathbf P s)=\mathbf P\mathbf Q(s).
\label{eq:permutation_main}
\end{equation}
Thus, the order-opening probability and value estimate are permutation
invariant, while the quantity vector is permutation equivariant.

\paragraph{Pathwise training.}
The order-opening decision $Y_t$ is discrete, whereas the quantities
$\mathbf Q_t$ are continuous. We train $Y_t$ with a score-based policy
gradient. Conditional on sampled order openings and demands, the
inventory dynamics are differentiable with respect to $\mathbf Q_t$; we
therefore backpropagate pathwise gradients through the resulting inventory
trajectory and future costs. Consequently, an order quantity at time $t$
receives gradients from later costs through its effects on future inventory
and outstanding orders. This yields more direct credit assignment for
high-dimensional quantity decisions than relying only on sampled returns.
A critic estimates expected future cost and provides a value estimate used
during training.

\section{Experiments}
\label{sec:experiments}

\paragraph{Experimental setting.}
We evaluate joint replenishment problems with
$n\in\{1,4,16,64,1{,}024\}$ items under correlated stochastic demand,
heterogeneous lead times, heterogeneous item costs, and a
shared setup cost. We compare OR-Transformer with three learning baselines. Transformer-PPO uses the Transformer backbone with PPO policy gradients for the quantity decisions; HPO uses a two-hidden-layer multilayer perceptron (MLP) with pathwise gradients \citep{alvo2026policy_arxiv,madeka2022deep_arxiv}; and PPO \citep{schulman2017proximal_arxiv} uses a two-hidden-layer MLP with PPO policy gradients.
We also compare against rolling-horizon
stochastic MILPs solved by Gurobi \citep{gurobi2026optimizer_manual} and HiGHS
\citep{huangfu2018parallelizing_mathematical_programming_computation}.
All learned methods use the same number of optimization updates and the same number of samples per update. Performance is evaluated on
a held-out set of 128 episodes using discounted inventory cost, where lower is
better. Full experimental specifications are provided in
Appendix~\ref{app:experimental_details}.

\begin{figure}[H]
\centering
\includegraphics[width=0.8\textwidth]{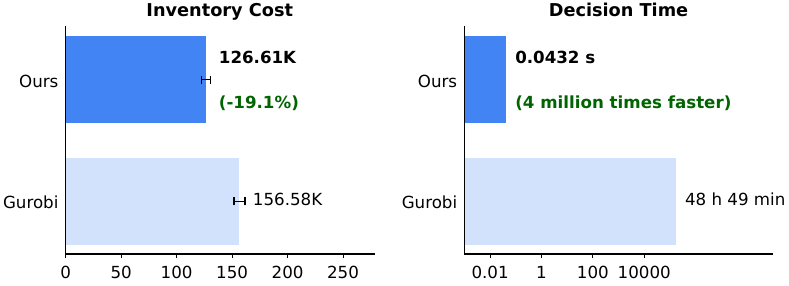}
\caption{\textbf{OR-Transformer versus Gurobi at 1,024 items.}
With Gurobi allowed up to six hours of optimization per decision,
OR-Transformer achieves 19.1\% lower inventory cost and over
\(4\times10^6\) times faster online decision-making on the same held-out
episodes. Error bars show one standard error; lower is better.}
\label{fig:ours_against_gurobi_eight_decisions}
\end{figure}

\paragraph{Scaling performance.}
Figure~\ref{fig:line_plots_row} shows that the baselines degrade at different scales as problem size grows. At 4 items, PPO is the first method to diverge. At 16 items, HPO also diverges, indicating that high-dimensional joint replenishment remains challenging for an MLP policy even with pathwise gradients. At 64 items, Transformer-PPO diverges as well, indicating that PPO policy gradients become increasingly difficult for the high-dimensional quantity decisions even with the Transformer architecture. OR-Transformer remains stable and scales to 1,024 items, where its cost falls below the rolling-horizon MILP baselines.
At 1,024 items, over a 50-decision evaluation horizon, OR-Transformer achieves approximately 0.35M discounted cost, compared with 1.39M for the best learning baseline, a reduction of approximately 75\%. 
Complete cost comparisons across all problem sizes, including rolling-horizon MILP results, are reported in Appendix~\ref{app:cost_results}.

\paragraph{Online decision quality and computation.}
We further compare OR-Transformer with rolling-horizon Gurobi at 1,024 items
over an eight-decision evaluation horizon, while allowing Gurobi up to
\textit{six hours} of optimization for each replenishment decision.
As shown in Figure~\ref{fig:ours_against_gurobi_eight_decisions},
OR-Transformer achieves \(126.61\pm4.23\)K discounted cost, compared with
\(156.58\pm4.93\)K for Gurobi, corresponding to a \textbf{19.1\% cost reduction}.
The OR-Transformer decisions require 0.0432 seconds in total, compared with 48 hours 49 minutes for Gurobi, making OR-Transformer more than \textbf{four million times faster} in online decision-making.
Full comparisons are reported in Appendix~\ref{app:milp_solver_setup}.

\section{Conclusion}

We introduced \textbf{OR-Transformer}, a deep RL framework for large-scale stochastic joint replenishment with an item-permutation-equivariant Transformer architecture and pathwise-gradient training through the inventory dynamics. Across problem sizes up to 1,024 items, OR-Transformer increasingly outperforms learning-based and rolling-horizon MILP baselines as scale grows, while reducing online decision-making time by more than 4 million times relative to MILP solvers. These results suggest a path toward efficient, real-time decision-making for large-scale stochastic inventory systems.

\section*{Acknowledgments}

David Simchi-Levi was supported in part by the Purdue University Center for Data Science. Shangtong Zhang was supported in part by the U.S. National Science Foundation under Awards III-2128019, SLES-2331904, and CAREER-2442098; the Commonwealth Cyber Initiative's Central Virginia Node under Award VV-1Q26-001; and a Cisco Faculty Research Award.

\clearpage

\bibliographystyle{apalike}
\bibliography{references}

@article{baek2026ai,
  title={AI agents for inventory control: Human-LLM-OR complementarity},
  author={Baek, Jackie and Fu, Yaopeng and Ma, Will and Peng, Tianyi},
  journal={arXiv preprint arXiv:2602.12631},
  year={2026}
}

@article{xie2024vc,
  title={Vc theory for inventory policies},
  author={Xie, Yaqi and Ma, Will and Xin, Linwei},
  journal={arXiv preprint arXiv:2404.11509},
  year={2024}
}

@article{xie2026deepstock,
  title={Deepstock: Reinforcement learning with policy regularizations for inventory management},
  author={Xie, Yaqi and Hao, Xinru and Liu, Jiaxi and Ma, Will and Xin, Linwei and Cao, Lei and Zhang, Yidong},
  journal={arXiv preprint arXiv:2603.19621},
  year={2026}
}

@article{chen2026robust,
  title={Robust Data-Collection Policy Learning for Low-Variance Online Policy Evaluation},
  author={Chen, Claire and Liu, Shuze Daniel and Luo, Licheng and Chandra, Rohan and Jiang, Nan and Zhang, Shangtong},
  journal={arXiv preprint arXiv:2608.24146},
  year={2026}
}

@inproceedings{chen2024efficient,
  title={Efficient Policy Evaluation with Safety Constraint for Reinforcement Learning},
  author={Chen, Claire and Liu, Shuze and Zhang, Shangtong},
  booktitle={Proceedings of the International Conference on Learning Representations},
  year={2025}
}

@inproceedings{liu2024doubly,
  title={Doubly Optimal Policy Evaluation for Reinforcement Learning},
  author={Liu, Shuze and Chen, Claire and Zhang, Shangtong},
  booktitle={Proceedings of the International Conference on Learning Representations},
  year={2025}
}

@inproceedings{liu2024efficient,
  author={Liu, Shuze and Zhang, Shangtong},
  booktitle={Proceedings of the International Conference on Machine Learning},
  title={Efficient Policy Evaluation with Offline Data Informed Behavior Policy Design},
  year={2024}
}

@inproceedings{liu2024efficientmul,
  title={Efficient Multi-Policy Evaluation for Reinforcement Learning},
  author={Liu, Shuze and Chen, Yuxin and Zhang, Shangtong},
  booktitle={Proceedings of the AAAI Conference on Artificial Intelligence},
  year={2025}
}

@misc{liu2026pessimistic,
  title  = {Pessimistic Minimax Learning for Public-Private Information Games under Unilateral Coverage},
  author = {Liu, Shuze Daniel and Chen, Claire and Wang, Jiuqi and Simchi-Levi, David},
  year   = {2026},
  note   = {Manuscript}
}

@article{wang2026predicting,
  title={Predicting Plasticity in Deep Continual Learning: A Theoretical Perspective},
  author={Wang, Jiuqi and Srinivasa, Jayanth and Chen, Claire and Liu, Shuze Daniel and Payani, Ali and Zhang, Shangtong},
  journal={arXiv preprint arXiv:2605.09044},
  year={2026}
}

@article{zhang2026pessimism,
      title={Beyond Pessimism: Offline Learning in KL-regularized Games}, 
      author={Yuheng Zhang and Claire Chen and Nan Jiang},
        journal={arXiv preprint arXiv:2604.06738},
  year={2026}
}

@inproceedings{chen2026offline,
title={Offline Two-Player Zero-Sum Markov Games with {KL} Regularization},
author={Claire Chen and Yuheng Zhang and Xinyu Liu and Zixuan Xie and Shuze Daniel Liu and Nan Jiang},
booktitle={Forty-third International Conference on Machine Learning},
year={2026},
url={https://openreview.net/forum?id=cQD2wxFhFG}
}

@article{chen2026pessimism,
  title={Pessimism-Free Offline Learning in General-Sum Games via {KL} Regularization},
  author={Chen, Claire and Zhang, Yuheng},
  journal={ArXiv Preprint arXiv:2605.00264},
  year={2026}
}

@article{chen2026fast,
  title={Fast Rates in $\alpha$-Potential Games via Regularized Mirror Descent},
  author={Chen, Claire and Zhang, Yuheng},
  journal={ArXiv Preprint arXiv:2605.00268},
  year={2026}
}

@phdthesis{liu2025efficientrobust,
  title     = {Efficient and Robust Policy Evaluation for Reinforcement Learning},
  author    = {Liu, Shuze Daniel},
  school    = {University of Virginia},
  year      = {2025}
}

@article{xie2026beyond,
  title={Beyond Linear Attention: {Softmax} Transformers Implement In-Context Reinforcement Learning},
  author={Xie, Zixuan and Liu, Xinyu and Chen, Claire and Liu, Shuze Daniel and Chandra, Rohan and Zhang, Shangtong},
  journal={arXiv preprint arXiv:2605.07333},
  year={2026}
}

@inproceedings{liu2021optimal,
  title={Optimal Pricing of Information},
  author={Liu, Shuze and Shen, Weiran and Xu, Haifeng},
  booktitle={Proceedings of the 22nd ACM Conference on Economics and Computation},
  pages={693},
  year={2021},
  publisher={ACM}
}

@article{liu2026strategic,
  title={Strategic Bargaining in Multi-Buyer Markets: Reinforcement Learning from Verifiable Rewards for {LLM} Negotiations},
  author={Liu, Shuze Daniel and Chen, Claire and Xiao, Jiabao Sean and Chen, Xin and Simchi-Levi, David},
  journal={arXiv preprint arXiv:2607.05863},
  year={2026}
}

@inproceedings{mahadevan2026convergence,
  title={Convergence of Two-Timescale {Markovian} Stochastic Approximations with Applications in Reinforcement Learning},
  author={Mahadevan, Vagul and Chen, Claire and Liu, Shuze Daniel and Zhang, Shangtong},
  booktitle={Proceedings of the 43rd International Conference on Machine Learning},
  year={2026}
}

@article{liu2026instructing,
  title={Instructing {LLM}s to Negotiate Using Reinforcement Learning with Verifiable Rewards},
  author={Liu, Shuze Daniel and Chen, Claire and Xiao, Jiabao Sean and Lei, Lei and Zhang, Yuheng and Yue, Yisong and Simchi-Levi, David},
  journal={arXiv preprint arXiv:2604.09855},
  year={2026}
}

@article{liu2025ode,
  title={The ODE method for stochastic approximation and reinforcement learning with markovian noise},
  author={Liu, Shuze Daniel and Chen, Shuhang and Zhang, Shangtong},
  journal={Journal of Machine Learning Research},
  volume={26},
  number={24},
  pages={1--76},
  year={2025}
}

@inproceedings{schulman2015gradient_neurips,
  author = {John Schulman and Nicolas Heess and Theophane Weber and Pieter Abbeel},
  title = {Gradient Estimation Using Stochastic Computation Graphs},
  booktitle = {Conference on Neural Information Processing Systems (NeurIPS)},
  year = {2015}
}

@article{alvo2026policy_arxiv,
  author = {Matias Alvo and Daniel Russo and Yash Kanoria},
  title = {Policy Optimization in Hybrid Discrete-Continuous Action Spaces via Mixed Gradients},
  journal = {arXiv preprint arXiv:2605.14297},
  year = {2026}
}

@article{schulman2017proximal_arxiv,
  author = {John Schulman and Filip Wolski and Prafulla Dhariwal and Alec Radford and Oleg Klimov},
  title = {Proximal Policy Optimization Algorithms},
  journal = {arXiv preprint arXiv:1707.06347},
  year = {2017}
}

@inproceedings{haarnoja2018soft_icml,
  author = {Tuomas Haarnoja and Aurick Zhou and Pieter Abbeel and Sergey Levine},
  title = {Soft Actor-Critic: Off-Policy Maximum Entropy Deep Reinforcement Learning with a Stochastic Actor},
  booktitle = {International Conference on Machine Learning (ICML)},
  year = {2018}
}

@article{khouja2008review_ejor,
  author = {Khouja, Moutaz and Goyal, Suresh},
  title = {A review of the joint replenishment problem literature: 1989–2005},
  journal = {European Journal of Operational Research (EJOR)},
  year = {2008}
}

@article{liu2023efficient_mathematics,
  author = {Liu, Shiyu and Liu, Ou and Jiang, Xiaoming},
  title = {An Efficient Algorithm for the Joint Replenishment Problem with Quantity Discounts, Minimum Order Quantity and Transport Capacity Constraints},
  journal = {Mathematics},
  year = {2023}
}

@article{goyal1974determination_management_science,
  author = {Goyal, S. K.},
  title = {Determination of Optimum Packaging Frequency of Items Jointly Replenished},
  journal = {Management Science},
  year = {1974}
}

@misc{ihl2026inventory_report,
  author = {{IHL Group}},
  title = {{The 2026 Inventory Distortion Study}},
  howpublished = {\url{https://www.ihlservices.com/product/inventory-distortion-study-2026/}},
  year = {2026}
}

@inproceedings{vaswani2017attention_neurips,
  author = {Ashish Vaswani and Noam Shazeer and Niki Parmar and Jakob Uszkoreit and Llion Jones and Aidan N. Gomez and Lukasz Kaiser and Illia Polosukhin},
  title = {Attention is All you Need},
  booktitle = {Conference on Neural Information Processing Systems (NeurIPS)},
  year = {2017}
}

@inproceedings{lee2019set_icml,
  author = {Juho Lee and Yoonho Lee and Jungtaek Kim and Adam R. Kosiorek and Seungjin Choi and Yee Whye Teh},
  title = {Set Transformer: {A} Framework for Attention-based Permutation-Invariant Neural Networks},
  booktitle = {International Conference on Machine Learning (ICML)},
  year = {2019}
}

@article{xiang2019rs_arxiv,
  author = {Mengyuan Xiang and Roberto Rossi and S. Armagan Tarim},
  title = {An $(R, S)$ Based Heuristic Model for the Stochastic Joint Replenishment Problem},
  journal = {arXiv preprint arXiv:1902.11025},
  year = {2019}
}

@article{chen2023robust_msom,
  author = {Chen, Yupeng and Iyengar, Garud and Wang, Chun},
  title = {Robust Inventory Management: A Cycle-Based Approach},
  journal = {Manufacturing and Service Operations Management (M\&SOM)},
  year = {2023}
}

@article{vanvuchelen2020ppo_computers_industry,
  author = {Vanvuchelen, Nathalie and Gijsbrechts, Joren and Boute, Robert},
  title = {Use of Proximal Policy Optimization for the Joint Replenishment Problem},
  journal = {Computers in Industry},
  year = {2020}
}

@article{madeka2022deep_arxiv,
  author = {Dhruv Madeka and Kari Torkkola and Carson Eisenach and Anna Luo and Dean P. Foster and Sham M. Kakade},
  title = {Deep Inventory Management},
  journal = {arXiv preprint arXiv:2210.03137},
  year = {2022}
}

@article{liu2026inventory_transformer_service_science,
  author = {Liu, Mo and Bai, Yumo and Qi, Meng and Shen, Zuo-Jun (Max)},
  title = {Inventory Management with Transformer: Automated Decision Making for Order Timing and Quantity},
  journal = {Service Science},
  year = {2026}
}

@inproceedings{zhang2024wims_iclr,
  author = {Chuheng Zhang and Xiangsen Wang and Wei Jiang and Xianliang Yang and Siwei Wang and Lei Song and Jiang Bian},
  title = {Whittle Index with Multiple Actions and State Constraint for Inventory Management},
  booktitle = {International Conference on Learning Representations (ICLR)},
  year = {2024}
}

@article{barati2026gym_arxiv,
  author = {Reza Barati and Qinmin Vivian Hu},
  title = {gym-invmgmt: An Open Benchmarking Framework for Inventory Management Methods},
  journal = {arXiv preprint arXiv:2605.11355},
  year = {2026}
}

@article{huangfu2018parallelizing_mathematical_programming_computation,
  author = {Qi Huangfu and J. A. J. Hall},
  title = {Parallelizing the dual revised simplex method},
  journal = {Mathematical Programming Computation},
  year = {2018}
}

@misc{gurobi2026optimizer_manual,
  author = {{Gurobi Optimization, LLC}},
  title = {{Gurobi Optimizer Reference Manual}},
  howpublished = {\url{https://www.gurobi.com}, self-published by the vendor, no publication year intrinsic to the document; retrieved 2026-08-30},
  year = {2026}
}
\clearpage
\appendix
\section{Notation}
\label{app:notation}

\begingroup
\small
\setlength{\tabcolsep}{4pt}
\renewcommand{\arraystretch}{1.10}
\begin{longtable}{@{}p{0.20\linewidth}p{0.23\linewidth}p{0.51\linewidth}@{}}
\toprule
Category & Symbol & Definition \\
\midrule
\endfirsthead
\toprule
Category & Symbol & Definition \\
\midrule
\endhead

Sets and indices
& $\mathcal I=\{1,\ldots,n\}$ & Item set containing $n$ items. \\
& $n$ & Number of items. \\
& $i,j$ & Item index and dummy item index used in symmetric sums. \\
& $t$ & Time step. \\
& $m$ & Temporal lag used in generalized advantage estimation. \\
& $\ell$ & Arrival offset associated with an in-transit quantity. \\
& $L_{\max}$ & Maximum admissible lead time. \\
& $B,e$ & Number of parallel rollouts and rollout index. \\
& $T_{\mathrm{roll}}$ & Number of time steps in one training rollout. \\
\midrule

Instance parameters
& $K$ & Common setup cost incurred when a joint order is opened. \\
& $\gamma$ & Infinite-horizon discount factor. \\
& $h_i,b_i$ & Per-unit holding cost and per-unit backlog cost of item $i$. \\
& $\lambda_i$ & Baseline demand rate of item $i$. \\
& $\beta_i$ & Loading of item $i$ on the common market factor. \\
& $\rho$ & Autoregressive coefficient of the market factor. \\
& $L_i$ & Deterministic lead time of item $i$. \\
& $\bar q_i$ & Order-quantity cap of item $i$. \\
& $\Theta$ & Collection of fixed parameters defining one problem instance. \\
\midrule

Demand and exogenous variables
& $F_t$ & Observed scalar market factor at time step $t$. \\
& $\varepsilon_t$ & Standard-normal innovation of the market factor. \\
& $\lambda_{i,t}$ & Conditional demand rate of item $i$ at time step $t$. \\
& $D_{i,t}$ & Realized demand of item $i$ at time step $t$. \\
\midrule

State
& $I_{i,t}$ & Net inventory of item $i$ immediately before demand at time step $t$; negative values represent backlog. \\
& $P_{i,\ell,t}$ & In-transit quantity of item $i$ that becomes available at time step $t+\ell$. \\
& $\mathbf I_t$ & Net-inventory vector at time step $t$. \\
& $\mathbf P_t$ & Array collecting all in-transit quantities at time step $t$. \\
& $s_t$ & Complete policy state at time step $t$. \\
& $s_0$ & Initial state. \\
\midrule

Action and dynamics
& $a_t$ & Hybrid replenishment action at time step $t$. \\
& $Y_t$ & Binary order-opening decision. \\
& $Q_{i,t},\mathbf Q_t$ & Proposed quantity of item $i$ and proposed-quantity vector. \\
& $O_{i,t}$ & Executed quantity, $O_{i,t}=Y_tQ_{i,t}$. \\
& $Z_{i,t}$ & Post-demand net inventory, $Z_{i,t}=I_{i,t}-D_{i,t}$. \\
& $\mathbf 1\{\cdot\}$ & Indicator function. \\
& $[x]^+$ & Positive part of scalar $x$, equal to $\max\{x,0\}$. \\
\midrule

Cost and control objective
& $C_t$ & One-step cost. \\
& $\pi$ & Stationary hybrid policy. \\
& $\mathbb E_\pi$ & Expectation under policy randomization and the exogenous processes. \\
& $J_\Theta(\pi;s_0)$ & Expected infinite-horizon discounted cost from initial state $s_0$. \\
\midrule

Feature representation
& $\mu_i$ & Lead-time-plus-one-step demand scale, $\mu_i=\lambda_i(L_i+1)$. \\
& $\sigma_i$ & Item reference scale, $\sigma_i=\sqrt{\mu_i}$. \\
& $c_{\mathrm{scale}}$ & Symmetric cost scale used in token normalization. \\
& $\sigma_{\mathrm{ref}}$ & Symmetric demand scale used in token normalization. \\
& $\mathbf x_{i,t}$ & Eleven-dimensional item token. \\
& $\mathbf g_t$ & Four-dimensional global token. \\
\midrule

Transformer architecture
& $r\in\{Y,Q,V\}$ & Network label for the order-opening network, quantity network, or critic. \\
& $d$ & Token width. \\
& $M$ & Number of Transformer blocks. \\
& $H_{\mathrm{attn}}$ & Number of attention heads. \\
& $E_G^r,E_I^r$ & Global-token embedding and shared item-token embedding of network $r$. \\
& $\mathcal T^r$ & Transformer encoder of network $r$. \\
& $\mathbf h_{0,t}^r$ & Final global representation produced by network $r$. \\
& $\mathbf H_t^r$ & Matrix of final item representations produced by network $r$. \\
& $\mathbf h_{i,t}^r$ & Final representation of item $i$ produced by network $r$. \\
& $\mathbf w_Y,c_Y$ & Output weight and bias of the order-opening head. \\
& $\mathbf w_Q,c_Q$ & Shared output weight and bias of the quantity head. \\
& $\mathbf w_V,c_V$ & Output weight and bias of the critic head. \\
& $\phi,\kappa,\psi$ & Parameter collections of the order-opening network, quantity network, and critic. \\
& $s,y,\mathbf q$ & Generic state, generic order-opening value, and generic proposed-quantity vector. \\
& $p_\phi(s),p_t$ & Order-opening probability function and its value at time step $t$. \\
& $V_\psi(s)$ & Normalized value-function estimate at state $s$. \\
& $\pi_\phi^Y(y\mid s)$ & Bernoulli order-opening distribution. \\
& $f_\kappa^Q(s)$ & Deterministic proposed-quantity vector. \\
& $\mathbf v,\delta_{\mathbf v}$ & Generic vector and point mass at that vector. \\
& $\pi_{\phi,\kappa}$ & Complete hybrid policy. \\
& $\tau,\mathbf P_\tau$ & Item permutation and its permutation matrix. \\
\midrule

Training
& $c_{\mathrm{ref}}$ & Fixed instance-level cost normalizer. \\
& $\widetilde C_t$ & Normalized one-step cost, $C_t/c_{\mathrm{ref}}$. \\
& $\lambda_{\mathrm{GAE}}$ & Trace parameter for generalized advantage estimation. \\
& $\delta_t$ & One-step temporal-difference residual in normalized-cost units. \\
& $\widehat A_t^C$ & Estimated cost advantage. \\
& $\widehat G_t$ & Bootstrapped target for the normalized value function. \\
& $\operatorname{sg}(\cdot)$ & Stop-gradient operator. \\
& $\bar\psi$ & Frozen critic parameters used in the update of the quantity network. \\
& $\mathcal L_Y$ & score-based loss of the order-opening network. \\
& $\mathcal L_{\mathrm{PW}}$ & Pathwise objective of the quantity network. \\
& $\mathcal L_V$ & Critic regression loss. \\
& $\mathcal H$ & Shannon entropy of the Bernoulli order-opening distribution. \\
& $\alpha_{\mathrm{PW}},\alpha_V,\alpha_H$ & Weights of the pathwise, critic, and entropy terms. \\
& $\mathcal L$ & Complete training objective. \\
\bottomrule
\end{longtable}
\endgroup

\section{Detailed Problem Formulation}
\label{app:problem_formulation}
This appendix provides the detailed formulation of the stochastic JRP summarized in Section~\ref{sec:problem}, including the demand process, state and action definitions, lead-time dynamics, and discounted-cost objective.

We study a stochastic joint replenishment problem (JRP), in which a common setup
cost couples the replenishment decisions of multiple items
\citep{goyal1974determination_management_science,khouja2008review_ejor}.
Let $n\in\mathbb N$ be the number of items, and define
$\mathcal I=\{1,\ldots,n\}$ as the item set. Let $i\in\mathcal I$ denote an
item, and let $t\in\mathbb N_0$ be a time step. For item $i$, denote its
per-unit holding cost, per-unit backlog cost, baseline demand rate, loading on
the common market factor, and order-quantity cap by
$h_i>0$, $b_i>0$, $\lambda_i>0$, $\beta_i\geq 0$, and $\bar q_i>0$,
respectively. Define $L_{\max}\in\mathbb N$ as the maximum admissible lead
time and $L_i\in\{1,\ldots,L_{\max}\}$ as the deterministic lead time of item
$i$.
We define $K \geq 0$ as the fixed joint setup cost incurred whenever an order is placed, regardless of the number of items assigned positive quantities. We consider an infinite-horizon planning model with a discount factor $\gamma \in (0,1)$. The system operates under backlogging, meaning any unmet demand is backlogged and fulfilled in subsequent periods rather than lost.

\paragraph{Correlated demand.}
Let $F_t\in\mathbb R$ denote the observed scalar market factor at time step
$t$, and let $\rho\in[0,1)$ be its autoregressive coefficient. Denote the
standard-normal noise by $\varepsilon_t\in\mathbb R$ and the initial
market factor by $F_0$. The factor process satisfies
\begin{equation}
  F_{t+1}
  =\rho F_t+\sqrt{1-\rho^2}\,\varepsilon_{t+1},
  \qquad
  F_0,\varepsilon_1,\varepsilon_2,\ldots
  \overset{\mathrm{i.i.d.}}{\sim}\mathcal N(0,1).
  \label{eq:factor_process}
\end{equation}

Let $\lambda_i$ be the base demand rate of item $i$. At time $t$, conditional on $F_t$, the realized demand $D_{i,t}$ for item $i$ satisfy

\begin{align}
 & D_{i,t}\mid F_t \sim\operatorname{Poisson}(\lambda_{i,t}), \label{eq:conditional_demand}  \\
  &\text{where}~ \lambda_{i,t}:=\lambda_i\exp\!\left(\beta_iF_t-\frac{\beta_i^2}{2}\right) \label{eq:conditional_demand_rate}.
\end{align}
We have
$\mathbb E[\lambda_{i,t}]=\lambda_i$. The factor process and demand
realizations are exogenous, i.e., their distributions do not depend on the action. Here, the market factor is observed without cost; extending the model to
endogenous acquisition or pricing of demand information would connect inventory
control with information-design problems \citep{liu2021optimal}.

\paragraph{State.}
Let $I_{i,t}\in\mathbb R$ denote the net inventory of item $i$ immediately
before demand at time step $t$; a negative value represents backlog. Let
$\ell\in\{1,\ldots,L_{\max}-1\}$ be an arrival offset, and let
$P_{i,\ell,t}\geq 0$ denote the in-transit quantity of item $i$ that becomes
available at time step $t+\ell$. Under this convention, the in-transit
quantity with arrival offset one becomes available only after demand and cost
at the current time step have been processed. Define
$\mathbf I_t=(I_{1,t},\ldots,I_{n,t})^\top\in\mathbb R^n$ as the net-inventory
vector and
$\mathbf P_t=(P_{i,\ell,t})_{i\in\mathcal I,\,1\leq\ell<L_{\max}}
\in\mathbb R_+^{n\times(L_{\max}-1)}$ as the array collecting all in-transit
quantities. We also define
\begin{equation}
  \Theta
  =\left(
    K,\gamma,\rho,
    \{h_i,b_i,\lambda_i,\beta_i,L_i,\bar q_i\}_{i=1}^{n}
  \right)
  \label{eq:instance_parameters}
\end{equation}
as the collection of fixed parameters defining one problem instance, and
\begin{equation}
  s_t=(\mathbf I_t,\mathbf P_t,F_t;\Theta)
  \label{eq:state}
\end{equation}
as the state observed at time step $t$. The parameters in $\Theta$ remain
fixed along a trajectory but are included in the policy input so that one
policy can act across heterogeneous instances.

\paragraph{Action and within-step timing.}
Let $Y_t\in\{0,1\}$ be the binary order-opening decision at time step $t$,
let $Q_{i,t}\in[0,\bar q_i]$ denote the \textit{proposed} quantity of item $i$, and
define $\mathbf Q_t=(Q_{1,t},\ldots,Q_{n,t})^\top$ as the proposed-quantity
vector. Denote the hybrid replenishment action $a_t$ selected at the beginning of
time step $t$ by
\begin{equation}
  a_t=(Y_t,\mathbf Q_t),
  \qquad
  Y_t\in\{0,1\},
  \qquad
  0\leq Q_{i,t}\leq\bar q_i.
  \label{eq:hybrid_action}
\end{equation}
We also define $O_{i,t}=Y_tQ_{i,t}$ as the \textit{executed} quantity of item $i$.
After the action is selected, demand is realized. Let
$Z_{i,t}$ denote the post-demand net inventory, defined by
\begin{equation}
  Z_{i,t}=I_{i,t}-D_{i,t}.
  \label{eq:post_demand}
\end{equation}
Holding, backlog, and setup costs are then assessed before any due in-transit
quantity becomes available.

\paragraph{Inventory dynamics.}
Recall $P_{i,\ell,t}$ as the in-transit quantity of item $i$ at time $t$ which is scheduled to become available
at time $t+\ell$, and recall $O_{i,t}$ as the executed quantity. Let
$\mathbf 1\{\cdot\}$ be the indicator function. We use $P_{i,L_{\max},t}\equiv 0$ as a boundary convention. After
the current cost is assessed, the in-transit quantity with arrival offset one
becomes available, each remaining in-transit quantity moves to the next
smaller arrival offset, and the new order is assigned the arrival offset
implied by $L_i$. The transition satisfies
\begin{align}
  I_{i,t+1}
  &=Z_{i,t}+P_{i,1,t}+\mathbf 1\{L_i=1\}O_{i,t},
  \label{eq:inventory_transition}\\
  P_{i,\ell,t+1}
  &=P_{i,\ell+1,t}
    +\mathbf 1\{\ell=L_i-1\}O_{i,t},
  \qquad \ell=1,\ldots,L_{\max}-1.
  \label{eq:in_transit_transition}
\end{align}
Consequently, an order placed at time step $t$ first becomes available for
demand at time step $t+L_i$.

\paragraph{Cost and control objective.}
For any scalar $x\in\mathbb R$, let $[x]^+=\max\{x,0\}$ denote its positive
part. Recall $Z_{i,t}$ as the post-demand net inventory. Define $C_t$ as the
one-step cost,
\begin{equation}
  C_t
  =KY_t+
  \sum_{i=1}^{n}
  \left(
    h_i[Z_{i,t}]^+
    +b_i[-Z_{i,t}]^+
  \right).
  \label{eq:one_step_cost}
\end{equation}
Let $\pi$ be a stationary hybrid policy that maps a state to a distribution
over the binary order-opening decision and a feasible proposed-quantity
vector, and let $s_0$ denote the initial state. Denote expectation over policy
randomization, market-factor innovations, and demand realizations under
$\pi$ by $\mathbb E_\pi$. We define
\begin{equation}
  J_{\Theta}(\pi;s_0)
  :=\mathbb E_{\pi}\!\left[
    \left.
    \sum_{t=0}^{\infty}\gamma^t C_t
    \right|s_0,\Theta
  \right]
  \label{eq:policy_objective}
\end{equation}
as the expected infinite-horizon discounted cost. The control problem is
\begin{equation}
  \min_{\pi}\;J_{\Theta}(\pi;s_0)
  \quad
  \text{subject to
  \eqref{eq:factor_process}--\eqref{eq:in_transit_transition}.}
  \label{eq:control_objective}
\end{equation}
The resulting model combines a discontinuous global order-opening decision, a
high-dimensional continuous quantity decision, correlated stochastic demand,
and delayed item-specific action effects. More broadly, structured reinforcement-learning problems also arise beyond inventory
control, including automated negotiation and multi-buyer bargaining
\citep{liu2026instructing,liu2026strategic}.

\section{Detailed OR-Transformer Method}
\label{app:detailed_method}

This appendix provides additional details for OR-Transformer, summarized in
Section~\ref{sec:method}, including the state representation, the
item-permutation-equivariant Transformer architecture, and the training
objectives. The architecture uses separate networks for the discrete
order-opening decision, the continuous order quantities, and the critic,
allowing the two action components to be trained with different gradient
estimators.

\subsection{Symmetry-preserving state representation}
\label{sec:method_features}

Recall $\mathcal I$ as the item set and $n$ as the number of items. Also
recall $I_{i,t}$ as net inventory, $P_{i,\ell,t}$ as an in-transit quantity,
$\lambda_i$ as the baseline demand rate, $\beta_i$ as the market-factor
loading, $L_i$ as the deterministic lead time, $h_i$ and $b_i$ as the holding
and backlog costs, and $\bar q_i$ as the order-quantity cap of item $i$. Use
$j\in\mathcal I$ as a dummy item in symmetric averages. Define $\mu_i$ as a
lead-time-plus-one-step demand scale, $\sigma_i$ as its square-root reference
scale, $c_{\mathrm{scale}}$ as the symmetric cost scale, and
$\sigma_{\mathrm{ref}}$ as the symmetric demand scale:
\begin{equation}
  \mu_i=\lambda_i(L_i+1),
  \qquad
  \sigma_i=\sqrt{\mu_i},
  \qquad
  c_{\mathrm{scale}}
  =\frac{1}{n}\sum_{j=1}^{n}(h_j+b_j),
  \qquad
  \sigma_{\mathrm{ref}}
  =\frac{1}{n}\sum_{j=1}^{n}\sigma_j.
  \label{eq:feature_scales}
\end{equation}
Recall $L_{\max}$ as the maximum admissible lead time. The reported
architecture fixes $L_{\max}=4$. Let
$\mathbf x_{i,t}\in\mathbb R^{11}$ denote the item token of item $i$ at time
step $t$:
\begin{equation}
\mathbf x_{i,t}=
\left[
  \frac{I_{i,t}-\mu_i}{\sigma_i},\;
  \frac{P_{i,1,t}}{\sigma_i},\;
  \frac{P_{i,2,t}}{\sigma_i},\;
  \frac{P_{i,3,t}}{\sigma_i},\;
  \frac{h_i}{c_{\mathrm{scale}}},\;
  \frac{b_i}{c_{\mathrm{scale}}},\;
  \log(1+\lambda_i),\;
  \frac{\sqrt{\lambda_i}}{\sigma_{\mathrm{ref}}},\;
  \frac{\bar q_i}{\sigma_i},\;
  \beta_i,\;
  L_i
\right]^{\!\top}.
\label{eq:item_features}
\end{equation}
Recall $K$ as the common setup cost, $\gamma$ as the discount factor, and $F_t$
as the observed market factor. We also define
$\mathbf g_t\in\mathbb R^4$ as the global token at time step $t$:
\begin{equation}
  \mathbf g_t=
  \left[
    \frac{K}{c_{\mathrm{scale}}\sigma_{\mathrm{ref}}},\;
    \gamma,\;
    \frac{1}{100(1-\gamma)},\;
    F_t
  \right]^{\!\top}.
  \label{eq:global_features}
\end{equation}
The numerical constant $100$ is a fixed feature-scaling constant. Every pooled
normalizer in \eqref{eq:feature_scales} is a symmetric function of the item
set. Consequently, permuting item indices permutes the item tokens and leaves
the global token unchanged.

\subsection{Item-permutation-compatible Transformer policy}
\label{sec:method_architecture}

We adapt full self-attention \citep{vaswani2017attention_neurips,xie2026beyond} to an
unordered item set, following the permutation-symmetric design principle of
set attention models \citep{lee2019set_icml}. Let $r\in\{Y,Q,V\}$ label the
order-opening network, quantity network, or critic, respectively, and let
$d\in\mathbb N$ be the token width. For each $r$, denote its global-token
embedding, shared item-token embedding, and Transformer encoder by
$E_G^r:\mathbb R^4\to\mathbb R^d$,
$E_I^r:\mathbb R^{11}\to\mathbb R^d$, and $\mathcal T^r$, respectively. Let
$\mathbf h_{0,t}^r\in\mathbb R^d$ be the final global representation and
$\mathbf H_t^r\in\mathbb R^{n\times d}$ be the matrix of final item
representations. We write $\mathbf h_{i,t}^r\in\mathbb R^d$ for row $i$ of
$\mathbf H_t^r$. The encoder computation is
\begin{equation}
  \begin{bmatrix}
    (\mathbf h_{0,t}^r)^\top\\
    \mathbf H_t^r
  \end{bmatrix}
  =\mathcal T^r\!\left(
    \begin{bmatrix}
      E_G^r(\mathbf g_t)^\top\\
      E_I^r(\mathbf x_{1,t})^\top\\
      \vdots\\
      E_I^r(\mathbf x_{n,t})^\top
    \end{bmatrix}
  \right),
  \qquad r\in\{Y,Q,V\}.
  \label{eq:three_transformers}
\end{equation}
The three encoders have disjoint parameters. They use full noncausal attention,
no positional embeddings, no item-index embeddings, and a mask only for
padded item slots. Let $M=4$ be the number of pre-normalized Transformer
blocks, and let $H_{\mathrm{attn}}=8$ be the number of attention heads in the reported
large-scale configuration. The same configuration uses $d=128$, a
feed-forward width of $4d$, and zero dropout.

Denote the complete parameter collections of the order-opening network,
quantity network, and critic by $\phi$, $\kappa$, and $\psi$, respectively.
Recall $s_t$ as the state at time step $t$ and $Y_t$ as the binary
order-opening decision. Let $s$ be a generic state. Let
$\mathbf w_Y\in\mathbb R^d$ and $c_Y\in\mathbb R$ be the output weight and bias of the order-opening head. Define $p_\phi(s)\in(0,1)$ as the
order-opening probability at state $s$, and write $p_t=p_\phi(s_t)$ for its
value at time step $t$:
\begin{equation}
  p_t
  =\operatorname{sigmoid}\!\left(
    \mathbf w_Y^\top\mathbf h_{0,t}^Y+c_Y
  \right),
  \qquad
  Y_t\sim\operatorname{Bernoulli}(p_t).
  \label{eq:order_head}
\end{equation}
Recall $Q_{i,t}$ as the proposed quantity and $\bar q_i$ as its cap. Let
$\mathbf w_Q\in\mathbb R^d$ and $c_Q\in\mathbb R$ denote the shared output
weight and bias of the quantity head. The deterministic quantity network
outputs
\begin{equation}
  Q_{i,t}
  =\bar q_i\,
  \operatorname{sigmoid}\!\left(
    \mathbf w_Q^\top\mathbf h_{i,t}^Q+c_Q
  \right).
  \label{eq:quantity_head}
\end{equation}
Let $\mathbf w_V\in\mathbb R^d$ and $c_V\in\mathbb R$ be the output weight
and bias of the critic head, and let $V_\psi(s_t)\in\mathbb R$ denote the critic output:
\begin{equation}
  V_\psi(s_t)
  =\mathbf w_V^\top\mathbf h_{0,t}^V+c_V.
  \label{eq:value_head}
\end{equation}

Let $y\in\{0,1\}$ be a generic order-opening value, and let
$\mathbf q\in\prod_{i=1}^{n}[0,\bar q_i]$ denote a generic proposed-quantity
vector. Denote the Bernoulli distribution produced by the order-opening
network by $\pi_\phi^Y(y\mid s)$ and the deterministic quantity vector
produced by the quantity network by $f_\kappa^Q(s)$. For a generic vector
$\mathbf v$, let $\delta_{\mathbf v}$ denote a point mass at $\mathbf v$. We
also define $\pi_{\phi,\kappa}$ as the complete hybrid policy, which factorizes
as
\begin{equation}
  \pi_{\phi,\kappa}(y,\mathbf q\mid s)
  =\pi_\phi^Y(y\mid s)\,
   \delta_{f_\kappa^Q(s)}(\mathbf q).
  \label{eq:policy_factorization}
\end{equation}

Let $\tau$ be an arbitrary permutation of the item indices, and denote its
permutation matrix by $\mathbf P_\tau\in\{0,1\}^{n\times n}$. We use
$\mathbf P_\tau s$ for the state obtained by applying $\mathbf P_\tau$ to
every item-indexed component of $s$ while leaving global components unchanged.
The architecture satisfies
\begin{equation}
  p_\phi(\mathbf P_\tau s)=p_\phi(s),
  \qquad
  V_\psi(\mathbf P_\tau s)=V_\psi(s),
  \qquad
  f_\kappa^Q(\mathbf P_\tau s)
  =\mathbf P_\tau f_\kappa^Q(s).
  \label{eq:permutation_property}
\end{equation}
Thus, the order-opening probability and $V_\psi$ are permutation invariant.
The proposed-quantity vector is permutation equivariant.

\subsection{Hybrid pathwise training}
\label{sec:method_training}

Recall $Y_t$ as the discrete order-opening decision and $\mathbf Q_t$ as the
continuous proposed-quantity vector. The sampled $Y_t$ prevents ordinary
end-to-end differentiation. Conditional on sampled order-opening decisions and
exogenous factor and demand realizations, however, the simulator is piecewise
differentiable in the proposed quantities. We therefore use a score-based
gradient for the order-opening network and a pathwise gradient for the quantity
network, treating each rollout as a stochastic computation graph
\citep{schulman2015gradient_neurips}. 

Recall $C_t$ as the one-step cost, $K$ as the setup cost, $h_i$ and $b_i$ as
the holding and backlog costs, and $\sigma_i$ as the item reference scale.
Define $c_{\mathrm{ref}}>0$ as the fixed instance-level cost normalizer and
$\widetilde C_t$ as the normalized one-step cost:
\begin{equation}
  c_{\mathrm{ref}}
  =K+\sum_{i=1}^{n}(h_i+b_i)\sigma_i,
  \qquad
  \widetilde C_t=\frac{C_t}{c_{\mathrm{ref}}}.
  \label{eq:cost_normalization}
\end{equation}
The critic is trained to estimate discounted future cost in these normalized-cost units.

Recall $\gamma$ as the discount factor. Let
$T_{\mathrm{roll}}\in\mathbb N$ be the number of time steps in one training
rollout, and let $m\in\mathbb N_0$ be a temporal lag used in generalized
advantage estimation. Let
$\lambda_{\mathrm{GAE}}\in[0,1]$ denote its trace parameter. Define
$\delta_t$ as the one-step temporal-difference residual in normalized-cost
units, $\widehat A_t^C$ as the estimated cost advantage, and $\widehat G_t$ as
the bootstrapped value target:
\begin{align}
  \delta_t
  &=\widetilde C_t
    +\gamma V_\psi(s_{t+1})
    -V_\psi(s_t),
  \label{eq:td_residual}\\
  \widehat A_t^C
  &=\sum_{m=0}^{T_{\mathrm{roll}}-1-t}
    (\gamma\lambda_{\mathrm{GAE}})^m\delta_{t+m},
  \label{eq:cost_advantage}\\
  \widehat G_t
  &=\widehat A_t^C+V_\psi(s_t).
  \label{eq:gae}
\end{align}
A positive $\widehat A_t^C$ indicates that the estimated cost-to-go exceeds
the critic's baseline at $s_t$.

Let $B\in\mathbb N$ be the number of parallel rollouts in one update, and let
$e\in\{1,\ldots,B\}$ index a rollout. Before updating the order-opening
network, we standardize the cost advantages over all rollout--time pairs:
\begin{equation}
  \widehat A_{e,t}^{C,\mathrm{std}}
  =
  \frac{
    \widehat A_{e,t}^C
    -\operatorname{mean}_{e,t}\!\left(\widehat A_{e,t}^C\right)
  }{
    \operatorname{std}_{e,t}\!\left(\widehat A_{e,t}^C\right)+10^{-8}
  }.
  \label{eq:standardized_advantage}
\end{equation}
Here, the mean and standard deviation are computed over all
$BT_{\mathrm{roll}}$ rollout--time pairs. The critic target $\widehat G_{e,t}$
remains defined from the unstandardized advantage in \eqref{eq:gae}.

For any differentiable quantity $u$, let $\operatorname{sg}(u)$ denote the
stop-gradient operator, which preserves $u$ in the forward computation and
assigns it zero derivative in the backward computation. The order-opening
network is trained with the score-based loss
\begin{equation}
  \mathcal L_Y(\phi)
  =
  \frac{1}{BT_{\mathrm{roll}}}
  \sum_{e=1}^{B}
  \sum_{t=0}^{T_{\mathrm{roll}}-1}
  \log\pi_\phi^Y\!\left(
    Y_{e,t}\mid\operatorname{sg}(s_{e,t})
  \right)
  \operatorname{sg}\!\left(
    \widehat A_{e,t}^{C,\mathrm{std}}
  \right).
  \label{eq:discrete_loss}
\end{equation}
This loss uses uniform weighting across time steps. Because each rollout is
used for a single on-policy update, its gradient coincides with the initial
actor gradient of the PPO surrogate
\citep{schulman2017proximal_arxiv}.

For the quantity network, sampled order-opening decisions, demand realizations,
and factor innovations are fixed during differentiation, while gradients
propagate through every proposed quantity, inventory transition, and future
cost in the rollout. Let $\bar\psi=\operatorname{sg}(\psi)$ be a frozen copy
of the critic parameters used in the update of the quantity network. Denote
the pathwise objective of the quantity network by
$\mathcal L_{\mathrm{PW}}(\kappa)$:
\begin{equation}
  \mathcal L_{\mathrm{PW}}(\kappa)
  =\frac{1}{B}\sum_{e=1}^{B}
  \left[
    \sum_{t=0}^{T_{\mathrm{roll}}-1}
    \gamma^t\widetilde C_{e,t}
    +\gamma^{T_{\mathrm{roll}}}
     V_{\bar\psi}(s_{e,T_{\mathrm{roll}}})
  \right].
  \label{eq:pathwise_loss}
\end{equation}
The frozen terminal value approximates the normalized value beyond the
truncated rollout while preserving gradients through the terminal state into
$\kappa$.

Let $\mathcal L_V(\psi)$ denote the critic regression loss:
\begin{equation}
  \mathcal L_V(\psi)
  =\frac{1}{2BT_{\mathrm{roll}}}
  \sum_{e=1}^{B}
  \sum_{t=0}^{T_{\mathrm{roll}}-1}
  \left(
    V_\psi(\operatorname{sg}(s_{e,t}))
    -\operatorname{sg}(\widehat G_{e,t})
  \right)^2.
  \label{eq:critic_loss}
\end{equation}
Recall $s$ as a generic state. Let
$\mathcal H(\pi_\phi^Y(\cdot\mid s))$ denote the Shannon entropy of the
Bernoulli order-opening distribution at state $s$. Let
$\alpha_{\mathrm{PW}}\geq 0$, $\alpha_V\geq 0$, and $\alpha_H\geq 0$ be the
weights of the pathwise, critic, and entropy terms, respectively. We define
$\mathcal L$ as the complete training objective:
\begin{equation}
  \mathcal L
  =\mathcal L_Y
   +\alpha_{\mathrm{PW}}\mathcal L_{\mathrm{PW}}
   +\alpha_V\mathcal L_V
   -\frac{\alpha_H}{BT_{\mathrm{roll}}}
    \sum_{e=1}^{B}
    \sum_{t=0}^{T_{\mathrm{roll}}-1}
    \mathcal H\!\left(
      \pi_\phi^Y(\cdot\mid\operatorname{sg}(s_{e,t}))
    \right).
  \label{eq:total_loss}
\end{equation}
The entropy term regularizes the order-opening policy. More broadly, regularization
and pessimism have been used to control policy optimization in offline
two-player, general-sum, and potential games
\citep{zhang2026pessimism,chen2026offline,
chen2026pessimism,chen2026fast,liu2026pessimistic}.

After each update, the simulator trajectories continue, but the net-inventory
vector and the array containing the in-transit quantities are detached from
the previous computation graph. This implements truncated backpropagation
through time for a continuing, non-episodic control process \citep{liu2025ode, mahadevan2026convergence}.

\section{Experimental Details}
\label{app:experimental_details}

\subsection{Method, training, and baselines configuration}
\label{app:method_training_baselines_configuration}

\subsubsection{Method and training configuration}
\label{app:method_training_configuration}

OR-Transformer uses three independent Transformer
encoders: one for the order-opening network, one for the quantity network, and
one for the critic.  The discrete order-opening decision is trained with a
score-based estimator, whereas gradients for the continuous quantity vector
are propagated pathwise through the differentiable simulator.  Table~\ref{tab:app_method_training_configuration}
collects the settings needed to reproduce the reported model and training
procedure.

\begin{table}[H]
  \centering
  \footnotesize
  \setlength{\tabcolsep}{4pt}
  \renewcommand{\arraystretch}{1.12}
  \begin{tabular}{@{}L{0.30\linewidth}L{0.64\linewidth}@{}}
    \toprule
    Setting & Value \\
    \midrule
    State representation
      & 11 features per item token and 4 features in the global token;
        $L_{\max}=4$ and three in-transit entries per item. \\
    Transformer
      & 4 pre-normalized blocks, token width 128, 8 attention heads,
        feed-forward width 512, full noncausal attention, zero dropout, no
        query/key/value bias, and no positional or item-index embedding. \\
    Output heads
      & A Bernoulli order-opening probability from the global representation;
        deterministic capped item quantities from shared item heads; and a
        scalar normalized value from the critic's global representation. \\
    Gradient estimators
      & score-based gradient for the order-opening network; pathwise gradient through the
        simulator for the quantity network; squared regression loss for the
        critic.  \\
    Discount rate
      & Discount $\gamma=0.95$, GAE trace $\lambda_{\mathrm{GAE}}=0.96$. \\
    Loss weights
      & $\alpha_{\mathrm{PW}}=1.0$, $\alpha_V=0.13$; Bernoulli entropy weight
        annealed linearly from 0.01 to 0.001. \\
    Optimizer
      & Adam with stability constant $10^{-5}$; learning rate warmed linearly
        for 40 updates to $3\times10^{-4}$ and then cosine-decayed to
        $3\times10^{-5}$; gradient-norm clipping at 5.0 for each network. \\
    Rollout and batch
      & Rollout length 10 and 1,024 parallel rollouts, giving 10,240 simulator
        transitions per update. \\
    Training updates
      & 16,000 updates. \\
    \bottomrule
  \end{tabular}
  \caption{OR-Transformer architecture and training settings.}
  \label{tab:app_method_training_configuration}
\end{table}

\subsubsection{OR-Transformer training algorithm}
\label{app:or_transformer_training_algorithm}

Recall $B$ as the number of parallel rollouts in one update and
$T_{\mathrm{roll}}$ as the number of time steps in one training rollout
(Section~\ref{sec:method_training}), and define $N_{\mathrm{update}}$ as the
number of training updates.
Let $\operatorname{Step}(s,Y,\mathbf O)$ denote one simulator transition, including exogenous
sampling, cost evaluation by \eqref{eq:one_step_cost}, normalization by
\eqref{eq:cost_normalization}, and the dynamics
\eqref{eq:inventory_transition}--\eqref{eq:in_transit_transition}. During the
pathwise update, sampled order-opening decisions and exogenous realizations are
treated as constants.

\begin{algorithm}[H]
  \caption{OR-Transformer hybrid pathwise training}
  \label{alg:or_transformer_training}
  \begin{algorithmic}[1]
    \Require $B$, $T_{\mathrm{roll}}$, $N_{\mathrm{update}}$, $\gamma$,
             $\lambda_{\mathrm{GAE}}$, $\alpha_{\mathrm{PW}}$, $\alpha_V$,
             $\alpha_H$; initial parameters $(\phi,\kappa,\psi)$
    \State Initialize $\{s_{e,0}\}_{e=1}^{B}$
    \For{$j=1,\ldots,N_{\mathrm{update}}$}
      \For{$t=0,\ldots,T_{\mathrm{roll}}-1$}
        \ForAll{$e=1,\ldots,B$ \textbf{in parallel}}
          \State $\mathbf Q_{e,t}\gets f_\kappa^Q(s_{e,t})$, \quad
                 $Y_{e,t}\sim\pi_\phi^Y(\cdot\mid s_{e,t})$
          \State $\mathbf O_{e,t}\gets Y_{e,t}\mathbf Q_{e,t}$
          \State $(\widetilde C_{e,t},s_{e,t+1})
                 \gets\operatorname{Step}(s_{e,t},Y_{e,t},\mathbf O_{e,t})$
        \EndFor
      \EndFor
      \State Compute $\{\widehat A_{e,t}^C,\widehat G_{e,t}\}_{e,t}$ by
             \eqref{eq:td_residual}--\eqref{eq:gae}
      \State $\mathcal L_Y\gets$ one-update PPO surrogate of
             \eqref{eq:discrete_loss}
      \State $\bar\psi\gets\operatorname{sg}(\psi)$; compute
             $\mathcal L_{\mathrm{PW}}$ and $\mathcal L_V$ by
             \eqref{eq:pathwise_loss} and \eqref{eq:critic_loss}
      \State Form $\mathcal L$ by \eqref{eq:total_loss}
      \State Adam step on $(\phi,\kappa,\psi)$ with per-network clipping
      \State $s_{e,0}\gets
             \operatorname{Detach}_{\mathbf I,\mathbf P}
             (s_{e,T_{\mathrm{roll}}})$, $e=1,\ldots,B$
    \EndFor
  \end{algorithmic}
\end{algorithm}

\subsubsection{Baselines details}
\label{app:baselines_details}

Table~\ref{tab:app_baselines_details} lists the environment setup, the
baseline architectures, and the training and evaluation protocol for 
Figure~\ref{fig:line_plots_row}.
All methods are evaluated on the same fixed held-out set of 128 episodes, following the standard policy-evaluation setting of estimating performance \citep{liu2024doubly, liu2024efficient, liu2024efficientmul,  chen2024efficient, liu2025efficientrobust, chen2026robust}.

\begin{table}[H]
  \centering
  \scriptsize
  \setlength{\tabcolsep}{3.5pt}
  \renewcommand{\arraystretch}{1.04}
  \begin{tabular}{@{}L{0.30\linewidth}L{0.64\linewidth}@{}}
    \toprule
    Setting & Value \\
    \midrule
    Problem sizes
      & $n\in\{1,4,16,64,1{,}024\}$.  \\
    Shared setup
      & Setup cost $K=10n$  \\
    Correlated demand
      &  $F_{t+1}=0.8F_t+\sqrt{1-0.8^2}\,\varepsilon_{t+1}$, where $\varepsilon_{t+1}\sim\mathcal N(0,1)$; item demands are Poisson. 
\\
    Item heterogeneity
      & Over the 1{,}024
        items $h_i\in[0.6,2.0]$, $b_i\in[6.0,11.4]$ and
        $\lambda_i\in[2.0,7.0]$. No two items share the
        triple $(h_i,b_i,\lambda_i)$; 
        $\beta_i\sim\mathrm{Unif}[0.2,0.6]$ and
        $L_i\sim\mathrm{Unif}\{1,2,3,4\}$.  The order cap
        is $4\lambda_i$. \\
    Perceptron baselines
      & Two hidden layers of width 512 with hyperbolic-tangent activations,
        applied to the concatenated $11n+4$ input. \\
    Unified training budget
      & 1,024 parallel environments, rollout length 10, and 16,000 updates for
        every method and size. \\
    Training seeds
      & 30 different seeds at 1, 4, 16, and 64 items. One training run for 1,024 items due to expensive computational cost. \\
    Curve statistic
      & Ten-step discounted inventory management cost
        $\sum_{t=0}^{9}\gamma^t C_t$ against training update; both axes are
        logarithmic. \\
    Smoothing and binning
      & Ten-update moving average followed by 24 logarithmically spaced bins. \\
    Displayed uncertainty
      & For $n\leq64$, 30 seeds are trained per algorithm, and the shaded area is one
        standard error across them. At $n=1{,}024$, the shaded area is the 10th--90th percentile within each
        bin of the single run. \\
    Solver reference lines
      & Gurobi and HiGHS inventory management cost over the first 10 decisions of the fixed
        128-episode evaluation set, using the solver configuration in
        Table~\ref{tab:app_milp_solver_setup}. \\
    \bottomrule
  \end{tabular}
  \caption{Environment setup and the
baseline architectures for
  Figure~\ref{fig:line_plots_row}.}
  \label{tab:app_baselines_details}
\end{table}

\subsection{MILP solver setup}
\label{app:milp_solver_setup}

At each time step, Gurobi and HiGHS implement the same rolling-horizon
policy by solving the same mixed-integer linear
program (MILP). Recall $\mathcal I=\{1,\ldots,n\}$ as the item set,
$s_t$ as the observed state, $I_{i,t}$ as the net inventory of item $i$
immediately before demand at time step $t$, and $P_{i,\ell,t}$ as the
in-transit quantity that becomes available at time step $t+\ell$. Also recall
$F_t$ as the observed market factor; $Y_t$ as the joint order-opening decision;
$O_{i,t}$ as the executed quantity; $K$ as the common setup cost; $h_i$ and
$b_i$ as the holding and backlog costs; $L_i$ and $L_{\max}$ as the item lead
time and maximum lead time; $\bar q_i$ as the order cap; and $\gamma$ as the
discount factor. 

\paragraph{Scenario construction.}
At time step $t$, condition on the observed state $s_t$. Define
$H\in\mathbb N$ as the MILP planning horizon and
$\mathcal U=\{0,\ldots,H-1\}$ as the set of look-ahead steps. The local index
$u\in\mathcal U$ is reset to zero at each MILP solve, with $u=0$ representing
the current decision. Let $\Omega=\{1,\ldots,S\}$ denote the set of $S$
sampled demand scenarios, and set $p_\omega=1/S$ as the weight of scenario
$\omega\in\Omega$.

For each scenario $\omega$, define $F_0^\omega=F_t$ as the observed market
factor at the root. Recall $\rho$ as the factor autocorrelation, $\lambda_i$
as the baseline demand rate of item $i$, and $\beta_i$ as its factor loading.
Let $\varepsilon_{u+1}^\omega\sim\mathcal N(0,1)$ be the factor innovation at
look-ahead step $u+1$, sampled independently across scenarios and look-ahead steps. The
sampled factor path then follows
\begin{equation}
  F_{u+1}^\omega
  =\rho F_u^\omega+\sqrt{1-\rho^2}\,\varepsilon_{u+1}^\omega,
  \qquad u=0,\ldots,H-2.
  \label{eq:milp_factor_scenarios}
\end{equation}
Denote the sampled demand of item $i$ at look-ahead step $u$ in scenario $\omega$ by
$D_{i,u}^\omega$. Conditional on each sampled factor path, item demands are
independent and are generated from the same demand model as the environment:
\begin{equation}
  D_{i,u}^\omega\mid F_u^\omega
  \sim \operatorname{Poisson}\!\left(
    \lambda_i\exp\!\left(\beta_iF_u^\omega-\frac{\beta_i^2}{2}\right)
  \right),
  \qquad i\in\mathcal I,\ u\in\mathcal U,\ \omega\in\Omega.
  \label{eq:milp_demand_scenarios}
\end{equation}
Once sampled, $F_u^\omega$ and $D_{i,u}^\omega$ are data, rather than decision
variables, in the deterministic-equivalent MILP.

\paragraph{Decision variables.}
Define $I_{i,u}^\omega\in\mathbb R$ as the net inventory immediately
before demand at look-ahead step $u$ in scenario $\omega$. Let
$Y_u^\omega\in\{0,1\}$ denote the scenario-$\omega$ order-opening decision,
and let $O_{i,u}^\omega\geq0$ be the corresponding executed quantity. We also
define $R_{i,u}^{+,\omega}\geq0$ and $R_{i,u}^{-,\omega}\geq0$ as auxiliary
variables for, respectively, on-hand inventory and backlog after demand. Define
$Y_0\in\{0,1\}$ and $O_{i,0}\geq0$ as the shared root opening decision and
shared root executed quantity. Use the aliases
$Y_0^\omega=Y_0$ and $O_{i,0}^\omega=O_{i,0}$ for every
$\omega\in\Omega$. Future variables $Y_u^\omega$ and $O_{i,u}^\omega$ with
$u\geq1$ are scenario specific.

\paragraph{Rolling horizon MILP.}
For notational convenience, extend the observed in-transit state by setting
$P_{i,\ell,t}=0$ whenever $\ell\notin\{1,\ldots,L_{\max}-1\}$, and set
$O_{i,v}^\omega=0$ for every negative local index $v<0$. Define
$L_{\min}=\min_{i\in\mathcal I}L_i$ as the shortest item lead time. The
rolling-horizon optimization problem is
\begin{align}
  \min\quad
  &\sum_{\omega\in\Omega}p_\omega
    \sum_{u=0}^{H-1}\gamma^u
    \left[
      K Y_u^\omega
      +\sum_{i\in\mathcal I}
       \left(h_iR_{i,u}^{+,\omega}+b_iR_{i,u}^{-,\omega}\right)
    \right]
  \label{eq:milp_objective}\\
  \text{s.t.}\quad
  &I_{i,0}^\omega=I_{i,t},
  &&i\in\mathcal I,\ \omega\in\Omega,
  \label{eq:milp_initial_inventory}\\
  &R_{i,u}^{+,\omega}\geq I_{i,u}^\omega-D_{i,u}^\omega,
  &&i\in\mathcal I,\ u\in\mathcal U,\ \omega\in\Omega,
  \label{eq:milp_holding_linearization}\\
  &R_{i,u}^{-,\omega}\geq D_{i,u}^\omega-I_{i,u}^\omega,
  &&i\in\mathcal I,\ u\in\mathcal U,\ \omega\in\Omega,
  \label{eq:milp_backlog_linearization}\\
  &I_{i,u+1}^\omega
    =I_{i,u}^\omega-D_{i,u}^\omega
     +P_{i,u+1,t}+O_{i,u+1-L_i}^\omega,
  &&\begin{aligned}
      i&\in\mathcal I,\ \omega\in\Omega,\\[-2pt]
      u&=0,\ldots,H-2,
    \end{aligned}
  \label{eq:milp_inventory_dynamics}\\
  &0\leq O_{i,u}^\omega\leq\bar q_iY_u^\omega,
  &&i\in\mathcal I,\ u\in\mathcal U,\ \omega\in\Omega,
  \label{eq:milp_joint_order_coupling}\\
  &Y_0^\omega=Y_0,
  &&\omega\in\Omega,
  \label{eq:milp_root_opening_nonanticipativity}\\
  &O_{i,0}^\omega=O_{i,0},
  &&i\in\mathcal I,\ \omega\in\Omega,
  \label{eq:milp_root_quantity_nonanticipativity}\\
  &O_{i,u}^\omega=0,
  &&\begin{aligned}
      i&\in\mathcal I,\ \omega\in\Omega,\\[-2pt]
      u&=H-L_i,\ldots,H-1,
    \end{aligned}
  \label{eq:milp_quantity_tail_fixing}\\
  &Y_u^\omega=0,
  &&\omega\in\Omega,\quad u=H-L_{\min},\ldots,H-1,
  \label{eq:milp_opening_tail_fixing}\\
  &I_{i,u}^\omega\in\mathbb R,\quad
   R_{i,u}^{+,\omega},R_{i,u}^{-,\omega}\geq0,\quad
   Y_u^\omega\in\{0,1\},
  &&i\in\mathcal I,\ u\in\mathcal U,\ \omega\in\Omega.
  \label{eq:milp_domains}
\end{align}

Constraints~\eqref{eq:milp_holding_linearization}--\eqref{eq:milp_backlog_linearization}
represent the positive and negative parts of post-demand net inventory exactly:
because $h_i>0$ and $b_i>0$, both inequalities are tight at an optimum.
Equation~\eqref{eq:milp_inventory_dynamics} is the environment dynamics.
The term $P_{i,u+1,t}$ delivers an order in transit, while
$O_{i,u+1-L_i}^\omega$ makes an order placed at look-ahead step $u+1-L_i$ available
after its item-specific lead time. Constraint~\eqref{eq:milp_joint_order_coupling}
forces every executed quantity to zero when the common order is closed and
retains the item cap when it is open.
Constraints~\eqref{eq:milp_root_opening_nonanticipativity}--\eqref{eq:milp_root_quantity_nonanticipativity}
share the root action.
The implemented
controller is feasible because it executes only the common root action.

\paragraph{Notation used only in the MILP setup.}
For the operating procedure below, define $H_{\mathrm{dec}}$ as the number of
real time steps in an evaluation trajectory and $\tau_{\max}$ as the solver
time limit for one decision. We also define $\varepsilon_{\mathrm{MIP}}$ as the
requested relative MILP gap. Table~\ref{tab:milp_local_notation} summarizes all
local notation introduced in this subsection. 

\begin{table}[H]
  \centering
  \scriptsize
  \setlength{\tabcolsep}{4pt}
  \renewcommand{\arraystretch}{1.10}
  \begin{tabular}{@{}p{0.23\linewidth}p{0.69\linewidth}@{}}
    \toprule
    Symbol & Definition \\
    \midrule
    $H,\mathcal U,u$
      & MILP planning horizon, set $\mathcal U=\{0,\ldots,H-1\}$ of look-ahead
        steps, and one look-ahead step. \\
    $\Omega,S,\omega,p_\omega$
      & Scenario set, number of sampled scenarios, scenario index, and scenario
        weight; $p_\omega=1/S$. \\
    $F_u^\omega,\varepsilon_u^\omega,D_{i,u}^\omega$
      & Sampled market factor, standard-normal factor innovation, and sampled
        item demand at look-ahead step $u$ in scenario $\omega$. \\
    $I_{i,u}^\omega$
      & Net inventory immediately before demand sampling. \\
    $Y_u^\omega,O_{i,u}^\omega$
      & Scenario-indexed order-opening decision and executed quantity. \\
    $Y_0,O_{i,0}$
      & Shared root opening decision and shared root executed quantity. \\
    $R_{i,u}^{+,\omega},R_{i,u}^{-,\omega}$
      & Auxiliary nonnegative on-hand inventory and backlog variables after
        sampled demand. \\
    $L_{\min}$
      & Minimum item lead time, $L_{\min}=\min_{i\in\mathcal I}L_i$. \\
    $H_{\mathrm{dec}}$
      & Number of time steps over at evaluation. \\
    $\tau_{\max},\varepsilon_{\mathrm{MIP}}$
      & Per-decision solver time limit and requested relative MILP gap. \\
    \bottomrule
  \end{tabular}
  \caption{Notation introduced only for the rolling-horizon MILP.}
  \label{tab:milp_local_notation}
\end{table}

\paragraph{Rolling-horizon Algorithm.}
Recall $H_{\mathrm{dec}}$ as the evaluation horizon, $\tau_{\max}$ as the
per-decision time limit, and $\varepsilon_{\mathrm{MIP}}$ as the requested
relative gap. At every real time step, the model is rebuilt from the realized
state; scenario-specific future decisions from the preceding solve are
discarded. Algorithm~\ref{alg:rolling_milp} formalizes this operation.

\begin{algorithm}[H]
  \caption{Rolling-horizon MILP controller}
  \label{alg:rolling_milp}
  \begin{algorithmic}[1]
    \Require Initial state $s_0$; scenario count $S$; planning horizon $H$;
             decision horizon $H_{\mathrm{dec}}$; solver time limit
             $\tau_{\max}$; relative gap $\varepsilon_{\mathrm{MIP}}$
    \For{$t=0,\ldots,H_{\mathrm{dec}}-1$}
      \State Observe $I_{i,t}$, $P_{i,\ell,t}$, and $F_t$ for all relevant
             items and arrival offsets.
      \State Sample $S$ conditional factor-and-demand paths using
             \eqref{eq:milp_factor_scenarios}--\eqref{eq:milp_demand_scenarios}.
      \State Build \eqref{eq:milp_objective}--\eqref{eq:milp_domains} from the
             observed state and sampled paths.
      \State Supply the all-zero order plan as a feasible initial solution.
      \State Solve the MILP.
      \If{the solver returns a feasible incumbent}
        \State Set $Y_t=Y_0^\star$ and $O_{i,t}=O_{i,0}^\star$ for every
               item $i$, using the incumbent root action.
      \Else
        \State Define $Y_t=0$ and $O_{i,t}=0$ for every item $i$.
      \EndIf
      \State Execute only $(Y_t,(O_{i,t})_{i\in\mathcal I})$; discard all actions with $u\geq1$.
      \State Observe realized demand, advance the inventory system by one real
             time step, and retain the resulting in-transit quantities.
    \EndFor
  \end{algorithmic}
\end{algorithm}

\begin{table}[H]
\centering
\footnotesize
\setlength{\tabcolsep}{4pt}
\renewcommand{\arraystretch}{1.12}
\begin{tabular}{@{}p{0.30\linewidth}p{0.64\linewidth}@{}}
\toprule
Setting & Value \\
\midrule
Evaluation set
& The same fixed holdout set of 128 episodes is used for the neural
policies, Gurobi, and HiGHS. \\
Solver backends
& Gurobi and HiGHS. \\
Scenario approximation
& $S=100$ sampled demand scenarios per real decision. \\
Planning and execution
& Planning horizon $H=50$ time steps; only the shared root action is
executed, after which the MILP is rebuilt from the realized next state. \\
Target relative gap
& $\varepsilon_{\mathrm{MIP}}=0.005$ (0.5\%). \\
Standard solver budget
& $\tau_{\max}=600$ seconds per decision for
Figure~\ref{fig:line_plots_row} and Table~\ref{tab:cost_at_decision_horizon_50}. \\
Long-budget comparison
& $\tau_{\max}=21{,}600$ seconds (six hours) per decision for
Figure~\ref{fig:ours_against_gurobi_eight_decisions}. \\
\bottomrule
\end{tabular}
\caption{Rolling-horizon MILP solver setup.}
\label{tab:app_milp_solver_setup}
\end{table}

\subsection{Full scaling results}
\label{app:cost_results}

Table~\ref{tab:cost_at_decision_horizon_50} reports the complete cost
comparison across problem sizes over a 50-decision evaluation horizon.
The rolling-horizon MILP baselines use the 10-minute-per-decision solver
budget specified in Table~\ref{tab:app_milp_solver_setup}. HPO achieves the lowest cost at one
and four items, while OR-Transformer achieves the lowest cost from 16 items
onward. At 64 items, OR-Transformer reduces cost by approximately 20\%
relative to Gurobi. At 1,024 items, it reduces cost by approximately 75\%
relative to the best learned baseline and by 96\% relative to Gurobi.

Several learned baselines diverge as problem size increases (Figure~\ref{fig:line_plots_row}), highlighting the challenge of maintaining stable learning at scale. Representation plasticity has been studied as one factor associated with learning stability in deep networks over extended training \citep{wang2026predicting}.

\begin{table}[H]
  \centering
  \scriptsize
  \setlength{\tabcolsep}{2.4pt}
  \renewcommand{\arraystretch}{1.08}
  \begin{tabular}{@{}l|cccccc@{}}
    \toprule
    Items & Ours & Transformer-PPO & HPO & PPO & Gurobi & HiGHS \\
    \midrule
    1 & \underline{255.121} $\pm$ 0.482
      & 256.762 $\pm$ 0.394
      & \textbf{253.959} $\pm$ 0.118
      & 255.319 $\pm$ 0.098
      & 267.695 $\pm$ 3.730
      & 270.752 $\pm$ 3.966 \\
    4 & \underline{1{,}009.71} $\pm$ 1.12
      & 1{,}014.76 $\pm$ 0.58
      & \textbf{1{,}005.55} $\pm$ 0.22
      & 8{,}119.01 $\pm$ 896.58
      & 1{,}079.15 $\pm$ 13.17
      & 1{,}113.46 $\pm$ 13.69 \\
    16 & \textbf{4.495} $\pm$ 0.003K
       & 5.342 $\pm$ 0.376K
       & 31.133 $\pm$ 3.544K
       & 47.289 $\pm$ 2.348K
       & \underline{4.642} $\pm$ 0.094K
       & 5.749 $\pm$ 0.082K \\
    64 & \textbf{20.04} $\pm$ 0.01K
       & 123.78 $\pm$ 31.52K
       & 115.29 $\pm$ 7.21K
       & 254.86 $\pm$ 8.64K
       & \underline{24.96} $\pm$ 0.26K
       & 35.94 $\pm$ 0.86K \\
    1024 & \textbf{0.35} $\pm$ 0.01M
         & 3.84 $\pm$ 0.23M
         & 1.79 $\pm$ 0.09M
         & \underline{1.39} $\pm$ 0.08M
         & 9.35 $\pm$ 0.21M
         & 9.35 $\pm$ 0.21M \\
    \bottomrule
  \end{tabular}
  \caption{Mean discounted cost over 50 decisions, with one standard error.
  Lower is better; the best entry in each row is bold and the next best is
  underlined. K and M denote thousands and millions. The solver baselines use
  100 demand scenarios, a 50-period planning horizon, and a 10-minute limit
  per replenishment decision.}
  \label{tab:cost_at_decision_horizon_50}
\end{table}

For $n\leq64$, Gurobi achieves lower discounted cost than HiGHS whenever the
two solver results differ. At 1,024 items, however, the 100-scenario,
50-period MILP contains approximately 20 million columns, and all 6,400
evaluated decisions for both solvers reach the 10-minute time limit. Both
solvers return zero replenishment quantities at every decision, producing the
same inventory trajectory and therefore the same reported cost. Thus, the
identical 1,024-item solver results reflect the common no-order trajectory
under the time limit rather than convergence to a common optimized solution.

\end{document}